\documentclass[11pt]{article}
\usepackage[margin=1in]{geometry}
\usepackage{amsmath,amssymb,graphicx,url}
\usepackage[round]{natbib}
\usepackage[hidelinks]{hyperref}
\newenvironment{keywords}{\par\smallskip\noindent\textbf{Keywords:} }{\par}

\title{Physics-Informed Causal Reasoning for\\ Nonstationary Robot Dynamics}
\usepackage{times}
\usepackage{booktabs}
\usepackage{multirow}
\usepackage{array}
\usepackage{siunitx}

\author{%
  Jungjin Lee$^{1}$ \quad Se Jin Kim$^{1}$ \quad Sung Kwon On$^{2}$ \quad
  Insoon Yang$^{1,2}$\thanks{Corresponding author.}\\[6pt]
  \small $^{1}$Department of Electrical and Computer Engineering, Seoul National University, Seoul 08826, Korea\\
  \small $^{2}$Graduate School of AI, Seoul National University, Seoul 08826, Korea\\
  \small \texttt{\{jungbbal, kimsejin, sungkwon, insoonyang\}@snu.ac.kr}%
}
\date{}
\begin{document}

\maketitle

\begin{abstract}%

Reasoning about what an action will cause requires internalizing how a system behaves. As conditions seldom stay fixed, this
must often happen within a few interactions. Yet systems that look identical in a snapshot can respond differently
to the same action, depending on physical conditions the snapshot hides. Context-based models capture this
separation by summarizing recent transitions into a latent context, but do
not specify how that context should change the equations of motion. Physics-informed neural networks build those equations into the model, but
describe only a single system and cannot recognize when it has changed. We present Physics-Informed Causal Reasoning (PICR), which bridges the two.
Only a context, re-inferred from recent interactions, is updated; it
conditions a fixed physics-informed network that determines the robot's
inertia and potential energy, so the robot adapts to new physical conditions
from its own experience without retraining or a hand-specified robot model. On a Franka Panda arm in NVIDIA Isaac Sim, PICR predicts motion under changing and unseen payloads, reducing error on the latter by $1.6$--$3.8\times$ over baselines, and extends to counterfactual queries under alternative torques. Our analysis shows that the context, inferred solely from interaction, encodes payload physical properties, tracks their changes, and governs the arm's response. The gravity and inertia terms induced by the context further demonstrate
its physical validity, each agreeing with the simulator at the level of
individual Lagrangian components.

\end{abstract}

\begin{keywords}%
  Physics-informed learning, Causal reasoning, Nonstationary dynamics,
Context-based adaptation, Robotic manipulation.
\end{keywords}

\section{Introduction}

Human intelligence internalizes the world from brief experience. One
explanation, long held in cognitive science, is that it rests on an innate
a priori structure through which experience is interpreted~\citep{spelke2007core, lake2016building}.
Here, we focus on causal reasoning about how actions affect motion
through the underlying physical mechanisms~\citep{pearl2009causality,peters2017elements}.
We study nonstationary robot dynamics, where hidden physical conditions
change during operation and alter the torque--motion relationship.

A central form this structure has taken is a separation of the intrinsic
properties that define a system from the kinematic state in which it is
observed. The state is only a snapshot: two systems can look identical at a
given moment yet respond differently to the same action, because of
properties such as mass or friction that the snapshot does not
reveal~\citep{on2026physics,baradel2019cophy, scholkopf2021toward}. Hidden-parameter
MDPs~\citep{doshivelez2016hidden} and causal world
models~\citep{scholkopf2021toward, wang2022causal} make this separation
explicit, and context-based approaches~\citep{rakelly2019efficient, zintgraf2020varibad, lee2020context}
define the intrinsic component as a context vector inferred from a short
window of recent transitions. We adopt this structure, as it mirrors the
human case most directly: limited interaction yields a compact summary of
the system, from which the consequences of potential actions can be
anticipated.

Separating context from state does not specify how physical conditions
change the equations of motion. This matters for adaptation to
within-episode changes and generalization beyond the training range:
a payload change affects both inertial and gravitational responses.
Encoding these relationships provides a physical inductive bias for
context-based prediction~\citep{lutter2019deep,karniadakis2021physics}.
Physics-informed networks~\citep{greydanus2019hamiltonian,desai2021port,cranmer2020lagrangian},
in particular Lagrangian models~\citep{lutter2019deep}, provide this prior
knowledge, but they bind the law to a single system and cannot tell when that system has changed. CaDeLaC~\citep{schulze2025context} combines a
history-conditioned context with Lagrangian dynamics for adaptation, but presupposes a nominal robot model such as a URDF, taking its knowledge of the system as given rather than drawn from experience.

We introduce Physics-Informed Causal Reasoning (PICR) for
nonstationary robot dynamics. PICR infers a dynamics-relevant context
from recent interactions and uses Feature-wise Linear Modulation
(FiLM)~\citep{perez2018film} to condition shared Lagrangian networks.
The context modulates inertia and potential while preserving their
mechanical relationships for each fixed context. Network parameters
are learned offline; adaptation updates only the context from new
observations. Payload labels organize the context during training but
are not required at inference. The model thus exposes how inferred
context changes the mechanical quantities used to predict motion.

\begin{figure*}[!t]
    \centering
    \includegraphics[width=1\textwidth]{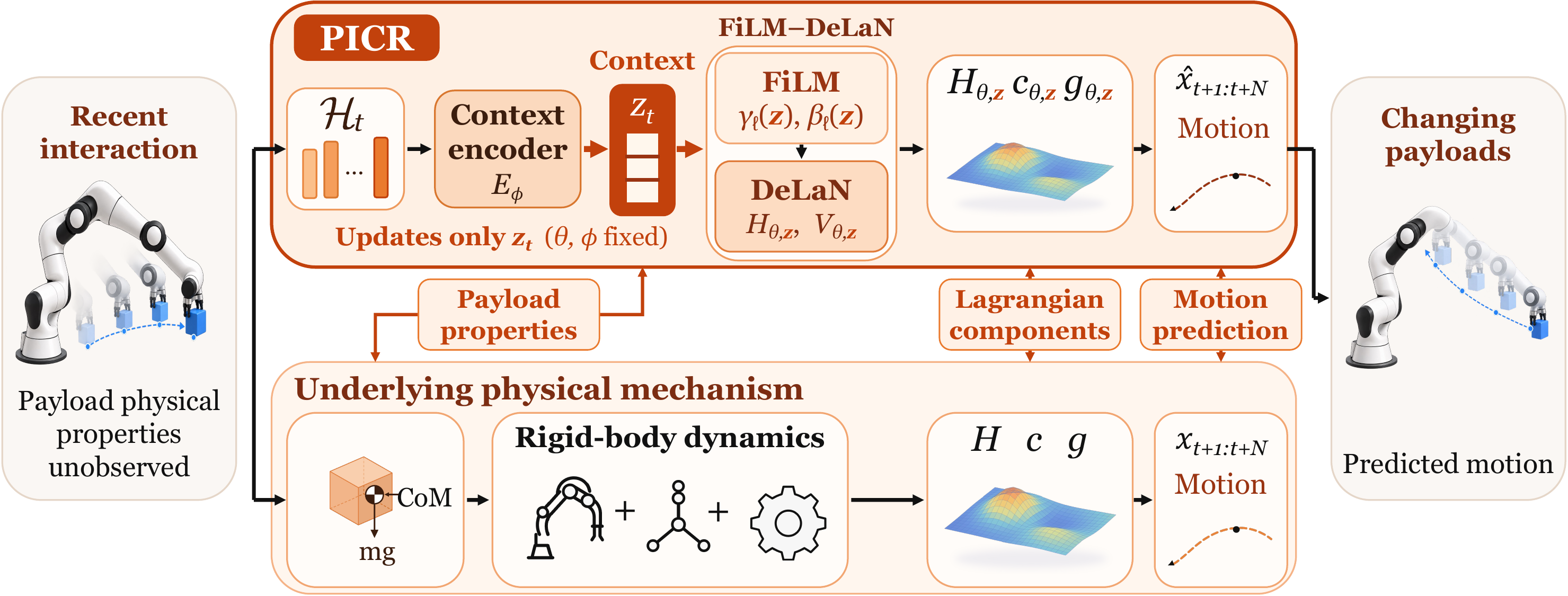}
\caption{PICR overview. A context inferred from recent interactions conditions a fixed Lagrangian network via FiLM, mirroring how unobserved payload properties shape the robot's rigid-body dynamics.}
    \label{fig:framework}
\end{figure*}

We evaluate PICR on a torque-controlled Franka Panda arm carrying a payload
of unknown mass and center of mass in NVIDIA Isaac Sim, against
context-based and physics-structured baselines. 


Our contributions are:
\begin{itemize}
\item A physics-informed causal reasoning framework for nonstationary
robot dynamics, in which interaction-derived context modulates a
shared Lagrangian model to adapt the modeled torque--motion
relationship while preserving its mechanical structure.

\item An empirical evaluation of within-episode adaptation,
generalization to unseen payloads, and action-effect prediction
through paired torque interventions.

\item A gray-box analysis linking inferred contexts, learned
mechanical terms, and motion predictions through context substitution
and comparison with simulator-derived references.

\end{itemize}
Our project page, including code, trained models, and videos, is available at \url{https://jungbbal.github.io/PICR/}.

\section{Related Works}

\paragraph{Causal reasoning}
A causal model predicts the effects of interventions and answers
counterfactual queries~\citep{pearl2009causality,peters2017elements,scholkopf2021toward}.
In dynamics learning, prior work models what changes across environments,
often through a latent context inferred from action--response
data~\citep{huang2022adarl,lee2020context,kumar2021rma,kirchmeyer2022generalizing}.
These methods largely leave the mechanism to a generic learned model, while
structured approaches~\citep{schulze2025context} and robot system
identification~\citep{wensing2018linear} rely on a
nominal robot model such as a URDF. We instead learn context-conditioned
robot--payload dynamics directly from interaction histories.

\paragraph{Physics-structured dynamics learning}
Physics-informed learning embeds physical knowledge into model
representations or equations of motion~\citep{karniadakis2021physics,raissi2019physics},
and mechanics-based networks derive motion from learned Hamiltonians or
Lagrangians~\citep{greydanus2019hamiltonian,cranmer2020lagrangian}.
We adopt the joint-space structure of DeLaN~\citep{lutter2019deep},
conditioned on context inferred from action--response history.

\section{Lagrangian Mechanics and Deep Lagrangian Networks}
\label{sec:preliminaries}

For a rigid-body manipulator with generalized coordinates
$q\in\mathbb R^n$~\citep{greenwood2006advanced, de2012theory, featherstone2007rigid},
the Lagrangian is $\mathcal L(q,\dot q)=\tfrac12\dot q^\top H(q)\dot q-V(q)$
with a symmetric positive definite inertia matrix $H(q)$.
Substituting $\mathcal L$ into the Euler--Lagrange equation with
non-conservative generalized forces $\tau$ yields the equations of motion
\begin{equation}
H(q)\,\ddot q
+ \underbrace{\dot H(q, \dot q)\,\dot q
 - \frac{1}{2}
   \left(\frac{\partial}{\partial q}\,
   \dot q^\top H(q)\,\dot q\right)^{\!\top}}_{c(q,\,\dot q)}
+ \underbrace{\nabla_q V(q)}_{g(q)}
 = \tau,
\label{eq:eom}
\end{equation}
where $c$ is the Coriolis and centrifugal force and $g$ the gravitational
force. Note that $c$ and $g$ are derived quantities, not independent
modeling components: the entire transition law is determined by $(H,V)$
alone.

The Deep Lagrangian Network (DeLaN)~\citep{lutter2019deep} incorporates this
structure directly into its parameterization. One network outputs a
lower-triangular matrix $A_\theta(q)$ with strictly positive diagonal
entries, and the other outputs a scalar potential $V_\theta(q)$, defining
$H_\theta(q)=A_\theta(q)A_\theta(q)^\top$; $c_\theta$ and $g_\theta$ then
follow from Eq.~\eqref{eq:eom}. This construction enforces the structural
properties of rigid-body dynamics, and the model is interpretable: trained
only on total torque, it still recovers the inertial, Coriolis--centrifugal,
and gravitational components individually.
A standard Lagrangian neural model, however, represents dynamics with fixed
parameters $\theta$ and therefore cannot adapt when the underlying dynamics
change over time. We address this limitation by inferring a context vector
from recent history and using it to condition the Lagrangian.

\section{Method}
\subsection{Problem formulation}
\label{sec:problem}

We study prediction of a robotic arm's joint-state
trajectory under unknown, potentially time-varying payload conditions. Given a recent
interaction history and a supplied future applied-torque sequence,
the task is to predict the resulting future joint states.
We consider a Franka Panda arm~\citep{haddadin2022franka}
carrying a cube in NVIDIA Isaac Sim~\citep{nvidia_isaacsim}
with Isaac Lab~\citep{mittal2023orbit}.
We restrict attention to the post-grasp regime, in which the cube
has been lifted and remains rigidly attached to the gripper
without slip. The gripper command is held constant, and the finger
joints are excluded from the prediction target.
The payload mass $m$ and gripper-frame center of mass $r$ are
unobserved at inference.
They are fixed within training episodes,
whereas evaluation includes within-episode mass changes.
Let $q_k,\dot q_k\in\mathbb{R}^7$ denote the arm joint positions and
velocities at step $k$, and $x_k=[q_k^\top,\dot q_k^\top]^\top\in\mathbb{R}^{14}$
the observed state. The torque input $\tau_k\in\mathbb{R}^7$ is the measured joint torque
averaged over $[k\Delta t,(k+1)\Delta t)$, with
$\Delta t=0.02\,\mathrm{s}$.

At prediction time $t\ge K$, the available history is
$\mathcal H_t=(x_{t-K+1:t},\,\tau_{t-K:t-1})$, where $a_{i:j}$ denotes
$(a_i,\ldots,a_j)$. We use $K=12$, so $\mathcal H_t$ contains $K-1$ complete transitions,
flattened into a 252-dimensional input. Let $F_\Theta$ denote a predictor with trainable parameters
$\Theta$. Given the history $\mathcal H_t$ and a supplied sequence of $N$ future applied torques, it produces
\begin{equation}
    \hat x_{t+1:t+N}
    =
    F_\Theta\!\left(
        \mathcal H_t,\tau_{t:t+N-1}
    \right),
    \qquad
    \hat x_t=x_t,
    \label{eq:predictor}
\end{equation}
where the hat
denotes a predicted state. The torque $\tau_t$, excluded from
$\mathcal H_t$, is the first control input for propagating $x_t$.

Payload properties, object observations, robot inertial parameters, and a
nominal dynamics model are not provided as inputs; payload labels are used
only during training (Section~\ref{sec:contrastive}).

\subsection{Context-conditioned Lagrangian dynamics}
\label{sec:model}

Physical conditions shape the system's equations of motion, which determine its
response to the current state and applied actions. Therefore, from a physical
standpoint, it is natural to view context as a factor governing the equations of
motion: $z$ determines a particular Lagrangian from a function family, and the
current state $(q,\dot q)$ is evaluated under that Lagrangian to determine the
response to an applied torque.

Let $z\in\mathbb R^4$ denote a context vector, whose value $z_t$ at prediction
time $t$ is inferred from the interaction history $\mathcal H_t$.
The context conditions the inertia matrix $H_{\theta,z}(q)\in\mathbb R^{7\times7}$
and the scalar potential $V_{\theta,z}(q)\in\mathbb R$, which define the Lagrangian
$L_{\theta,z}(q,\dot q)=\tfrac{1}{2}\dot q^\top H_{\theta,z}(q)\dot q-V_{\theta,z}(q)$
where $H_{\theta,z}(q)$ is positive definite and $\theta$ denotes the trainable
parameters, including the conditioning maps.
All derivatives with respect to $q$ are evaluated with $z$ held fixed.
Thus, context modifies the inertia and potential, while the velocity-dependent
and gravitational forces remain derived quantities.

\subsubsection{Context inference from interaction history}
\label{sec:context}

The interaction history records how the system has evolved under
the torques applied to it. We distill this evidence into a context
$z_t$ using an MLP encoder $E_\phi$ with trainable parameters
$\phi$, which maps the flattened history to the mean and variance of a
Gaussian $\mathcal N(\mu_t,\operatorname{diag}(\sigma_t^2))$
with $\mu_t,\sigma_t\in\mathbb R^4$.
During training, we sample $z_t=s(\mu_t+\sigma_t\odot\epsilon)$ with
$\epsilon\sim\mathcal N(0,I)$ and the elementwise soft bound
$s(x)=10\tanh(x/10)$ (Appendix~\ref{sec:app-encoder}), which allows the
dynamics losses to train the encoder jointly with the dynamics model; at
inference, we use $z_t=\mu_t$.

\subsubsection{Context-conditioned Lagrangian components}
\label{sec:conditioning}

We implement this conditioning by applying FiLM~\citep{perez2018film} to the
inertia-factor and potential networks of DeLaN (Section~\ref{sec:preliminaries}):
context-dependent scales and shifts modulate their hidden features, while the
backbone parameters remain shared across contexts. The shared layers process
$q$, while learned conditioning maps take $z$ and generate
layer-wise scales and shifts, transforming the hidden features
$\tilde h_\ell$ as
$h_\ell=\gamma_\ell(z)\odot\tilde h_\ell+\beta_\ell(z)$.
The resulting $A_{\theta,z}(q)$ and $V_{\theta,z}(q)$ define
$H_{\theta,z}$, $c_{\theta,z}$, and $g_{\theta,z}$ as in
Section~\ref{sec:preliminaries}, giving the conditioned dynamics
$\ddot q=H_{\theta,z}(q)^{-1}\bigl(\tau-c_{\theta,z}(q,\dot q)-g_{\theta,z}(q)\bigr)$. For trajectory prediction, we infer $z_t$ from $\mathcal H_t$
and initialize the rollout at the observed state $x_t$.
We perform an open-loop rollout under the supplied torques
$\tau_{t:t+N-1}$ using trapezoidal integration, without incorporating
future state observations and with $z_t$ fixed over the horizon. New observations update the context for subsequent
predictions; future context changes are not forecast.

\subsubsection{Physically structured contrastive learning}
\label{sec:contrastive}

Training serves two complementary purposes: fitting the observed
torque--motion relation and encouraging context to reflect physical
conditions consistently across different states and motions.
The dynamics losses make context useful for prediction, while the
contrastive objective encourages its organization according to
payload conditions.

\paragraph{Physics-structured dynamics loss}
For each training window, we infer one context from its history
and use it throughout the supervised transitions; all mechanical
terms are evaluated at the logged state using this context.
We compare the predicted acceleration with the acceleration
$\ddot q_{\mathrm{data}}$ estimated from recorded joint velocities,
$\mathcal L_{\ddot q}=\operatorname{MSE}\bigl(H_{\theta,z}^{-1}(\tau-c_{\theta,z}-g_{\theta,z}),\,\ddot q_{\mathrm{data}}\bigr)$,
and also fit the corresponding inverse-dynamics residual,
$\mathcal L_{\mathrm{ID}}=\operatorname{MSE}\bigl(H_{\theta,z}\ddot q_{\mathrm{data}}+c_{\theta,z}+g_{\theta,z},\,\tau\bigr)$,
where MSE is computed after per-joint scaling.
The two losses measure the same equation-of-motion discrepancy in
different units. Because $H_{\theta,z}$, $c_{\theta,z}$, and
$g_{\theta,z}$ are constructed from the conditioned Lagrangian, these
losses fit the data only through the equations of motion.

\begin{figure*}[!t]
    \centering
    \makebox[\textwidth][c]{%
        \includegraphics[width=1\textwidth]{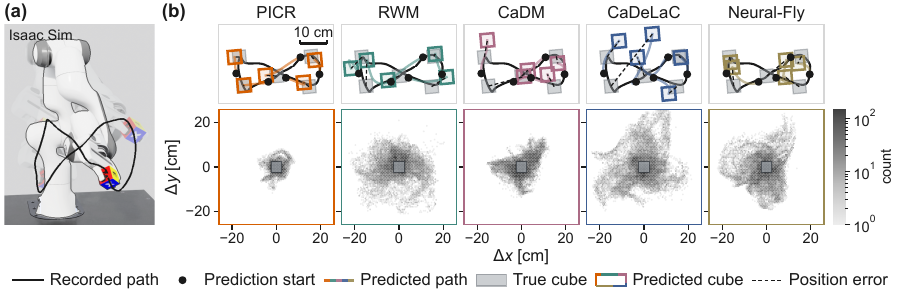}%
    }
\caption{\textbf{Prediction under a payload ramp (0.1$\to$4\,kg).}
(a) Figure-eight task in Isaac Sim. (b) Top: predicted (coloured) vs.\ true (grey) cube.
Bottom: 300\,ms center error over all windows; grey square: cube (4.8\,cm).}
    \label{fig:cubes-sweep}
\end{figure*}

\paragraph{Contrastive loss}
FiLM specifies how $z$ modulates the mechanics, but does not determine
what it contains. Under dynamics losses alone, $z$ can absorb
trajectory-specific features. We therefore use payload mass and
center-of-mass labels to define pairwise supervision, encouraging
context directions to be similar across different states and motions
observed under similar physical conditions. Together with the
Lagrangian structure, in which $z$ acts through $H$ and $V$, this
encourages context to capture the dynamical effects of payload
properties.
Each batch contains $M$ histories, with two temporally separated
histories from each sampled episode. For histories $i$ and $j$, we define
$w_{ij}=\exp\bigl(-(m_i-m_j)^2/2\sigma_m^2-\|r_i-r_j\|_2^2/2\sigma_r^2\bigr)$,
normalized per anchor as $\bar w_{ij}=w_{ij}/\sum_{k\ne i}w_{ik}$.
With $u_i=\mu_i/\|\mu_i\|_2$ and $s_{ij}=u_i^\top u_j/\kappa$,
\begin{equation}
    \mathcal L_{\mathrm{con}}
    =
    -\frac{1}{M}
    \sum_{i=1}^{M}\sum_{j\ne i}
    \bar w_{ij}
    \log
    \frac{\exp(s_{ij})}{\sum_{k\ne i}\exp(s_{ik})},
    \label{eq:contrastive}
\end{equation}
where $\kappa=0.1$ and all other histories remain in the softmax
denominator. The labels define only which histories should share
similar contexts; $z$ is never regressed onto $m$ or $r$.

The total objective is
$\mathcal L_{\mathrm{total}}=\lambda_a\mathcal L_{\ddot q}+\lambda_{\mathrm{ID}}\mathcal L_{\mathrm{ID}}+\lambda_{\mathrm{con}}\mathcal L_{\mathrm{con}}+\lambda_{\mathrm{KL}}\mathcal L_{\mathrm{KL}}$,
where $\mathcal L_{\mathrm{KL}}$ is the minibatch mean of
$D_{\mathrm{KL}}\!\left(
\mathcal N(\mu_t,\operatorname{diag}(\sigma_t^2))
\,\|\,\mathcal N(0,I)\right)$,
evaluated before applying the soft bound $s$.

\section{Experiments}
\label{sec:experiments}

We evaluate whether PICR predicts motion under changing and unseen payloads
(Sec.~\ref{sec:adapation_experiment}), answers counterfactual torque interventions
(Sec.~\ref{sec:counterfactual}), infers a context that reflects payload
properties (Sec.~\ref{sec:latent-physical}), and recovers the individual Lagrangian
torque components (Sec.~\ref{sec:dynamics-term-reconstruction}).
See Appendix~\ref{sec:app-impl} for data collection details and
Appendix~\ref{sec:app-model} for the model architecture.

\subsection{Experimental setup}

\paragraph{Setup.}
All experiments use Isaac Lab with PhysX (2\,kHz simulation, 50\,Hz control;
Appendix~\ref{sec:app-discrete-time-dynamics}). Training data come from joint-wise
cosine sweeps under PID control with Ornstein--Uhlenbeck torque perturbations, with
payload mass randomized over eight levels in 0.1--2.0\,kg and CoM over a $5^3$ lattice
(318 training / 40 validation episodes; Appendix~\ref{sec:app-impl}). Evaluation uses
PPO-generated figure-eight and carry trajectories. PICR uses mass and CoM labels during
training and Neural-Fly uses mass labels; label-supervised variants of the other
baselines are in Appendix~\ref{sec:app-supcon-baselines}.


\paragraph{Baseline.}
We compare PICR against five baselines: context-based adaptive models, \textbf{CaDM}~\citep{lee2020context}
and \textbf{RWM}~\citep{li2025robotic}, which infer dynamics from recent transitions
or history; physics-structured models, \textbf{DeLaN}~\citep{lutter2019deep} and its
context-conditioned extension \textbf{CaDeLaC}~\citep{schulze2025context}; and
\textbf{Neural-Fly}~\citep{o2022neural}, which adapts basis-function coefficients online.
Details are in Appendix~\ref{sec:baselines}.



\subsection{Causal dynamics inference and generalization}
\label{sec:causal-dynamics-inference}

\begin{figure}[t]
    \centering
    \includegraphics[width=\linewidth]{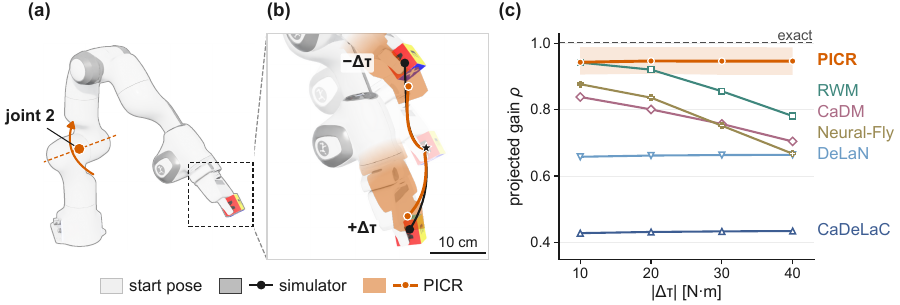}
\caption{\textbf{Counterfactual torque interventions.} (a) Intervention configuration.\
(b) Simulated vs.\ PICR motion 300\,ms after a $\pm$20\,N\,m offset to joint~2.
(c) Projected gain $\rho$ vs.\ $|\Delta\tau|$ (joint~2); shading: PICR IQR.}
    \label{fig:counterfactual-torque}
\end{figure}

\begin{table}[t]
    \centering
    \small
    \setlength{\tabcolsep}{4pt}
    \caption{Prediction, counterfactual and component evaluation (mean over three seeds). ADD: figure-eight payload sweep, seen (0.1--2\,kg) vs.\ unseen (2--4\,kg) range. Gain: median
    acceleration gain under $\pm$10--40\,N\,m torque offsets. RMSE: joint-2 torque
    terms vs.\ simulator on a 3\,kg carry. n/a: no explicit $H$, $c$, $g$. Best in bold.}
    \vspace{4pt} 
    \label{tab:causal-results}
    \begin{tabular}{@{}llcccccc@{}}
        \toprule
        Metric & Setting & PICR & RWM & CaDM & CaDeLaC & DeLaN & Neural-Fly \\
        \midrule
        \multirow{2}{*}{ADD (cm) $\downarrow$}
            & Seen (0.1--2\,kg)   & \textbf{4.93} & 7.45 & 6.06 & 8.75 & 6.96 & 8.05 \\
            & Unseen (2--4\,kg)   & \textbf{6.76} & 11.15 & 11.40 & 21.41 & 25.78 & 22.27 \\
        \midrule
        Gain $\rho$ $\rightarrow 1$
            & Joints 1--3 (mean)  & \textbf{0.90} & 0.75 & 0.65 & 0.17 & 0.35 & 0.84 \\
        \midrule
        \multirow{3}{*}{RMSE (N\,m) $\downarrow$}
            & Gravity $g_2$            & \textbf{2.15} & n/a & n/a & 11.29 & 9.99 & n/a \\
            & Inertial $[H\ddot{q}]_2$ & \textbf{2.31} & n/a & n/a & 12.49 & 5.83 & n/a \\
            & Coriolis $c_2$           & \textbf{1.59} & n/a & n/a & 2.15 & 1.67 & n/a \\
        \bottomrule
    \end{tabular}
\end{table}

\subsubsection{Generalization and adaptation across task and physical conditions}
\label{sec:adapation_experiment}

We first evaluate whether PICR can predict motion under changing physical conditions, including unseen payloads.
The predicted arm configuration is converted to cube pose using the grasp offset measured at the end of the history. We
report cube ADD~\citep{hinterstoisser2012model}, computed as the mean Euclidean displacement between
2,000 corresponding points on the ground-truth and predicted cube
surfaces; lower values indicate more accurate predictions.

We evaluate PICR as the arm tracks a figure-eight path while the payload mass increases
from 0.1 to 4.0\,kg; since the mass is not a model input, the current dynamics must be
inferred from recent interaction history. As shown in Table~\ref{tab:causal-results} and
Fig.~\ref{fig:cubes-sweep}, PICR attains the lowest ADD in both mass ranges despite the
within-episode change. The gap widens on payloads unseen in training: from the seen to the
unseen range, PICR's ADD increases by only 1.8\,cm, whereas that of CaDeLaC, DeLaN, and Neural-Fly increases by 12.7--18.8\,cm.

\subsubsection{Counterfactual prediction under torque interventions}
\label{sec:counterfactual}
PICR performs well across diverse states and under nonstationary payload changes. Accurate trajectory prediction, however, does not show whether a model captures how motion responds to changes in action, which is central to decision-making. To test action-effect prediction, we construct paired counterfactual runs that share the same pre-intervention history but receive different torque inputs. Each counterfactual run applies a torque offset of $\pm$10, 20, 30, or
40\,N\,m for 15 steps (300\,ms) to one of the first three joints, which have
the largest influence on the arm's motion and the payload trajectory
(Fig.~\ref{fig:counterfactual-torque}(a)).

For each counterfactual run, we measure the change in joint acceleration
relative to its paired reference run over the first 20\,ms after
intervention, denoted by $\Delta\ddot{q}_{\mathrm{sim}}$ for simulation (ground truth)
and $\Delta\ddot{q}_{\mathrm{pred}}$ for a model. We quantify the
predicted response magnitude along the simulated change 
using the projected gain
  $\rho=\langle \Delta\ddot q_{\mathrm{pred}},\Delta\ddot q_{\mathrm{sim}}\rangle/\|\Delta\ddot q_{\mathrm{sim}}\|_2^2$,
  where $\rho=1$ indicates a matched projected magnitude, not necessarily a matched
acceleration vector.

Fig.~\ref{fig:counterfactual-torque}(b) illustrates PICR's predicted motion under an intervention. In
Table~\ref{tab:causal-results}, PICR's joint-averaged gain is closest to 1, followed by Neural-Fly
and RWM, while the other baselines fall further below. Fig.~\ref{fig:counterfactual-torque}(c) shows
the gain across torque magnitudes: Neural-Fly and RWM lose gain as the offset grows,
whereas PICR's gain remains stable, indicating more consistent prediction of the response
along the simulated change.

\subsection{Latent representation of physical condition}
\label{sec:latent-physical}

\begin{figure*}[!t]
    \centering
    \makebox[\textwidth][c]{%
    \includegraphics[width=1\textwidth]{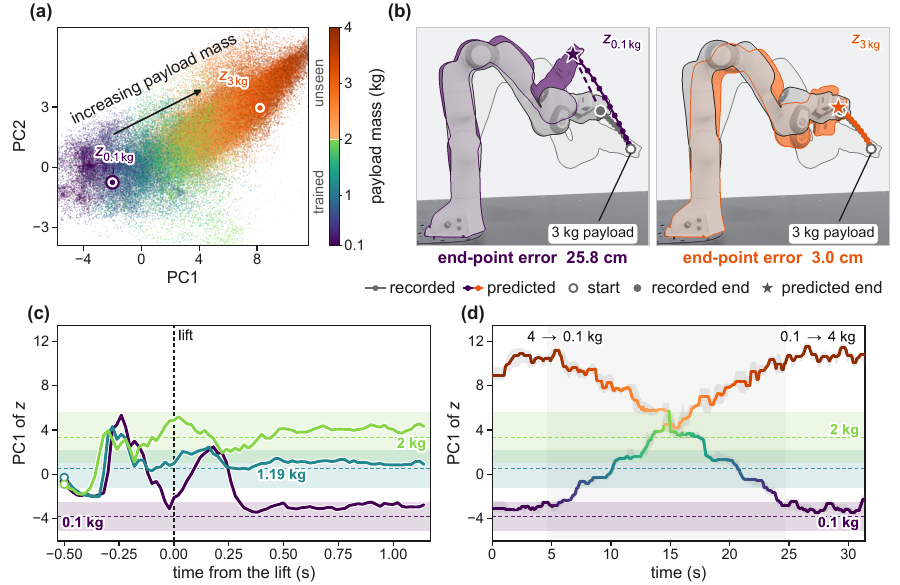}%
    }
\caption{\textbf{Physical information in the inferred context.} (a) PCA of $z$ coloured
by payload mass. (b) Context substitution on a 3\,kg window. (c) PC1 after payload lift.
(d) PC1 as the payload ramps between 0.1 and 4\,kg.}
    \label{fig:qualitative_ic}
\end{figure*}

We examine whether the four-dimensional context $z$, inferred online from state--torque history and conditioning the learned dynamics, reflects condition-dependent dynamics, how it evolves during interaction, and whether it affects predictions.

\paragraph{Context structure across physical conditions.}
\label{sec:context_structure}
To identify which physical properties \(z\) reflects, we apply PCA to \(z\) and examine its structure as payload mass and CoM vary. PC1 and PC2 capture 80\% of the variance in \(z\) and are organized primarily by mass. As shown in Fig.~\ref{fig:qualitative_ic}(a), PC1 increases with mass, and contexts from unseen heavier payloads extend along the same direction, indicating that this structure persists beyond the training range. In contrast, coloring the same projection by any CoM coordinate reveals no visible ordering (Fig.~\ref{fig:com-latent}). This is consistent with the training data, where CoM offsets produce only about 5\% of the mass-induced torque (Appendix~\ref{sec:app-com-grasp}). Regions for different masses partially overlap, likely because \(z\) is inferred solely from interaction history and thus varies with the visited states, applied torques, and task. How to handle this uncertainty, which arises from inferring physical conditions only implicitly, is left for future work.

\paragraph{Physical information in the context.}
Fig.~\ref{fig:qualitative_ic}(b) tests whether the dynamics model uses mass-related information in \(z\). For a 3\,kg evaluation window in which the robot carries an object of fixed mass, we fix the initial state and recorded future torques and vary only the context. The window's own context, \(z_{3\,\mathrm{kg}}\), yields a rollout close to the recorded trajectory. Substituting the mean context from 0.1\,kg episodes, \(z_{0.1\,\mathrm{kg}}\), makes the prediction drift upward \emph{as if the object were lighter}, raising the error to 25.8\,cm. Thus, $z$ carries mass-related information used by the dynamics model.

\paragraph{Context inferred from ongoing interaction.}
Fig.~\ref{fig:qualitative_ic}(c) tracks PC1 while the robot grasps an object and carries it to a fixed goal. Before the lift, contexts for different masses largely overlap, reflecting an uninformative initialization. After the lift, they separate by mass in real time, converging to the mass-dependent structure identified above. Fig.~\ref{fig:qualitative_ic}(d) extends this to a nonstationary setting. During figure-eight tracking (Sec.~\ref{sec:adapation_experiment}), as the payload mass increases from 0.1 to 4.0\,kg, PC1 moves into the heavier-payload region observed above, and it returns as the mass decreases. Thus, the context is inferred online, both at the onset of payload interaction and under continuous changes in the dynamics.

\subsection{Agreement of Lagrangian torque components}
\label{sec:dynamics-term-reconstruction}

\begin{figure*}[!t]
    \centering
    \makebox[\textwidth][c]{%
        \includegraphics[width=\textwidth]{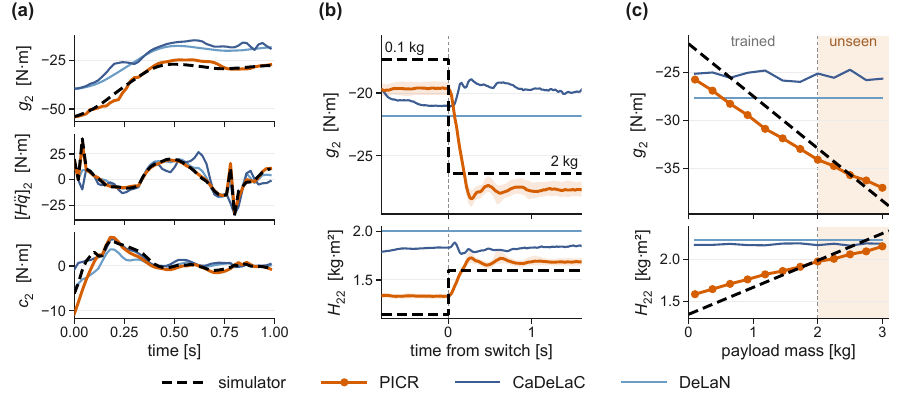}%
    }
\caption{\textbf{Learned Lagrangian terms follow the payload.} Joint-2 terms vs.\ simulator
(dashed). (a) 3\,kg carry (unseen). (b) Payload jump 0.1$\to$2\,kg (median of 472 episodes;
band: PICR IQR). (c) Across payload masses (shaded: unseen).}
    \label{fig:lagrangian-components}
\end{figure*}

Accurate state prediction alone does not show that the model captures the
physical contributions behind the robot's response. We therefore compare the
learned inertial \(H(q,z)\ddot q\), Coriolis \(c(q,\dot q,z)\), and gravity
\(g(q,z)\) torques with simulator references at each recorded state
(Appendix~\ref{sec:app-discrete-time-dynamics}).

First, Fig.~\ref{fig:lagrangian-components}(a) shows a fixed-payload carry with
an unseen 3\,kg payload. On joint 2, the shoulder joint that bears the largest
payload load, PICR's gravity and inertial errors are several times lower than
those of CaDeLaC and DeLaN (Table~\ref{tab:causal-results}), while all three
models perform similarly on the small Coriolis term. PICR thus shows closer agreement on gravity and inertia, not merely on their sum.

Next, Fig.~\ref{fig:lagrangian-components}(b) holds the robot state fixed while
the payload switches from 0.1 to 2.0\,kg during carry, so any change in the
predicted terms must come from the context. PICR's \(g_2\) and \(H_{22}\) reach
90\% of the reference change within 200\,ms, whereas those of CaDeLaC and DeLaN
barely move.
Finally, over a fixed bank of postures, Fig.~\ref{fig:lagrangian-components}(c)
shows that PICR follows the linear mass dependence of the simulator's \(g_2\)
and \(H_{22}\) and extends it into the unseen range, whereas CaDeLaC and DeLaN
stay nearly flat. The response is attenuated beyond the training range, so this agreement is
approximate rather than exact.

\section{Conclusion}
We presented PICR for nonstationary robot dynamics. Context inferred from recent interactions conditions shared Lagrangian networks, allowing adaptation without updating their parameters. Simulation results show prediction under changing and unseen payloads, responses to paired torque interventions conditional on measured torques, and approximate agreement with simulator-derived mechanical terms. The study is limited to a single simulated robot, rigid grasps, and open-loop prediction. Real-hardware evaluation, richer physical variation, and closed-loop control are natural next steps.

\section*{Acknowledgments}
This work was supported in part by 
the National Research Foundation of Korea (NRF) grants funded by MSIT No. RS-2026-25477173, No. RS-2026-25615715 and No. RS-2026-25504174, and the National IT Industry
Promotion Agency (NIPA) grant funded by MSIT No. RS-2026-25621677. Generative AI tools (ChatGPT by OpenAI and Claude by Anthropic) were used for writing assistance and code development; the authors reviewed all outputs and take full responsibility for the content of this paper.

\bibliographystyle{plainnat}
\bibliography{l4dc2027}

@article{wensing2018linear,
  title   = {Linear matrix inequalities for physically consistent inertial parameter
             identification: A statistical perspective on the mass distribution},
  author  = {Wensing, Patrick M. and Kim, Sangbae and Slotine, Jean-Jacques E.},
  journal = {IEEE Robotics and Automation Letters},
  volume  = {3},
  number  = {1},
  pages   = {60--67},
  year    = {2018}
}

@article{o2022neural,
  title     = {{Neural-Fly} enables rapid learning for agile flight in strong winds},
  author    = {O'Connell, Michael and Shi, Guanya and Shi, Xichen and Azizzadenesheli, Kamyar and Anandkumar, Anima and Yue, Yisong and Chung, Soon-Jo},
  journal   = {Science Robotics},
  volume    = {7},
  number    = {66},
  pages     = {eabm6597},
  year      = {2022},
  publisher = {American Association for the Advancement of Science}
}

@inproceedings{kumar2021rma,
  title     = {{RMA}: Rapid Motor Adaptation for Legged Robots},
  author    = {Kumar, Ashish and Fu, Zipeng and Pathak, Deepak and Malik, Jitendra},
  booktitle = {Robotics: Science and Systems (RSS)},
  year      = {2021},
  doi       = {10.15607/RSS.2021.XVII.011}
}

@inproceedings{doshivelez2016hidden,
  title     = {Hidden Parameter {Markov} Decision Processes: A Semiparametric Regression Approach for Discovering Latent Task Parametrizations},
  author    = {Doshi-Velez, Finale and Konidaris, George},
  booktitle = {Proceedings of the Twenty-Fifth International Joint Conference on Artificial Intelligence (IJCAI)},
  pages     = {1432--1439},
  year      = {2016}
}

@inproceedings{rakelly2019efficient,
  title     = {Efficient Off-Policy Meta-Reinforcement Learning via Probabilistic Context Variables},
  author    = {Rakelly, Kate and Zhou, Aurick and Finn, Chelsea and Levine, Sergey and Quillen, Deirdre},
  booktitle = {International Conference on Machine Learning (ICML)},
  pages     = {5331--5340},
  year      = {2019},
  organization = {PMLR}
}

@inproceedings{zintgraf2020varibad,
  title     = {{VariBAD}: A Very Good Method for {Bayes}-Adaptive Deep {RL} via Meta-Learning},
  author    = {Zintgraf, Luisa and Shiarlis, Kyriacos and Igl, Maximilian and Schulze, Sebastian and Gal, Yarin and Hofmann, Katja and Whiteson, Shimon},
  booktitle = {International Conference on Learning Representations (ICLR)},
  year      = {2020}
}

@inproceedings{lee2020context,
  title        = {Context-aware Dynamics Model for Generalization in Model-Based Reinforcement Learning},
  author       = {Lee, Kimin and Seo, Younggyo and Lee, Seunghyun and Lee, Honglak and Shin, Jinwoo},
  booktitle    = {International Conference on Machine Learning (ICML)},
  pages        = {5757--5766},
  year         = {2020},
  organization = {PMLR}
}

@inproceedings{kirchmeyer2022generalizing,
  title        = {Generalizing to new physical systems via context-informed dynamics model},
  author       = {Kirchmeyer, Matthieu and Yin, Yuan and Don{\`a}, J{\'e}r{\'e}mie and Baskiotis, Nicolas and Rakotomamonjy, Alain and Gallinari, Patrick},
  booktitle    = {International Conference on Machine Learning (ICML)},
  pages        = {11283--11301},
  year         = {2022},
  organization = {PMLR}
}

@inproceedings{perez2018film,
  title     = {{FiLM}: Visual reasoning with a general conditioning layer},
  author    = {Perez, Ethan and Strub, Florian and de Vries, Harm and Dumoulin, Vincent and Courville, Aaron},
  booktitle = {Proceedings of the AAAI Conference on Artificial Intelligence},
  volume    = {32},
  year      = {2018},
  doi       = {10.1609/aaai.v32i1.11671}
}

@book{pearl2009causality,
  title     = {Causality},
  author    = {Pearl, Judea},
  edition   = {2nd},
  year      = {2009},
  publisher = {Cambridge University Press}
}

@book{peters2017elements,
  title     = {Elements of Causal Inference: Foundations and Learning Algorithms},
  author    = {Peters, Jonas and Janzing, Dominik and Sch{\"o}lkopf, Bernhard},
  year      = {2017},
  publisher = {The MIT Press}
}

@article{scholkopf2021toward,
  title   = {Toward Causal Representation Learning},
  author  = {Sch{\"o}lkopf, Bernhard and Locatello, Francesco and Bauer, Stefan and Ke, Nan Rosemary and Kalchbrenner, Nal and Goyal, Anirudh and Bengio, Yoshua},
  journal = {Proceedings of the IEEE},
  volume  = {109},
  number  = {5},
  pages   = {612--634},
  year    = {2021}
}

@inproceedings{wang2022causal,
  title     = {Causal Dynamics Learning for Task-Independent State Abstraction},
  author    = {Wang, Zizhao and Xiao, Xuesu and Xu, Zifan and Zhu, Yuke and Stone, Peter},
  booktitle = {International Conference on Machine Learning (ICML)},
  pages     = {23151--23180},
  year      = {2022},
  organization = {PMLR}
}

@inproceedings{huang2022adarl,
  title     = {{AdaRL}: What, where, and how to adapt in transfer reinforcement learning},
  author    = {Huang, Biwei and Feng, Fan and Lu, Chaochao and Magliacane, Sara and Zhang, Kun},
  booktitle = {International Conference on Learning Representations (ICLR)},
  year      = {2022}
}

@inproceedings{baradel2019cophy,
  title     = {{CoPhy}: Counterfactual Learning of Physical Dynamics},
  author    = {Baradel, Fabien and Neverova, Natalia and Mille, Julien and Mori, Greg and Wolf, Christian},
  booktitle = {International Conference on Learning Representations (ICLR)},
  year      = {2020}
}

@article{on2026physics,
  title   = {Physics-informed causal reasoning in physical {AI}: A review on modeling non-stationary environments for safety-critical control},
  author  = {On, Sung Kwon and Jeong, Sukchul and Lee, Jungjin and Yang, Insoon},
  journal = {International Journal of Control, Automation, and Systems},
  volume  = {24},
  number  = {4},
  pages   = {726--752},
  year    = {2026},
  doi     = {10.1007/s12555-026-00082-x}
}

@article{spelke2007core,
  title     = {Core knowledge},
  author    = {Spelke, Elizabeth S. and Kinzler, Katherine D.},
  journal   = {Developmental Science},
  volume    = {10},
  number    = {1},
  pages     = {89--96},
  year      = {2007},
  publisher = {Wiley}
}

@article{lake2016building,
  title   = {Building machines that learn and think like people},
  author  = {Lake, Brenden M. and Ullman, Tomer D. and Tenenbaum, Joshua B. and Gershman, Samuel J.},
  journal = {Behavioral and Brain Sciences},
  volume  = {40},
  pages   = {e253},
  year    = {2017},
  doi     = {10.1017/S0140525X16001837}
}

@inproceedings{lutter2019deep,
  title     = {Deep {Lagrangian} Networks: Using Physics as Model Prior for Deep Learning},
  author    = {Lutter, Michael and Ritter, Christian and Peters, Jan},
  booktitle = {International Conference on Learning Representations (ICLR)},
  year      = {2019}
}

@inproceedings{cranmer2020lagrangian,
  title     = {{Lagrangian Neural Networks}},
  author    = {Cranmer, Miles and Greydanus, Sam and Hoyer, Stephan and Battaglia, Peter and Spergel, David and Ho, Shirley},
  booktitle = {ICLR 2020 Workshop on Integration of Deep Neural Models and Differential Equations},
  year      = {2020}
}

@inproceedings{greydanus2019hamiltonian,
  title     = {{Hamiltonian} neural networks},
  author    = {Greydanus, Samuel and Dzamba, Misko and Yosinski, Jason},
  booktitle = {Advances in Neural Information Processing Systems (NeurIPS)},
  volume    = {32},
  year      = {2019}
}

@article{desai2021port,
  title     = {Port-{Hamiltonian} neural networks for learning explicit time-dependent dynamical systems},
  author    = {Desai, Shaan A. and Mattheakis, Marios and Sondak, David and Protopapas, Pavlos and Roberts, Stephen J.},
  journal   = {Physical Review E},
  volume    = {104},
  number    = {3},
  pages     = {034312},
  year      = {2021},
  publisher = {APS}
}

@inproceedings{schulze2025context,
  title        = {Context-aware deep {Lagrangian} networks for model predictive control},
  author       = {Schulze, Lucas and Peters, Jan and Arenz, Oleg},
  booktitle    = {IEEE/RSJ International Conference on Intelligent Robots and Systems (IROS)},
  pages        = {6939--6946},
  year         = {2025},
  organization = {IEEE},
  doi          = {10.1109/IROS60139.2025.11246292}
}

@article{raissi2019physics,
  title     = {Physics-informed neural networks: A deep learning framework for solving forward and inverse problems involving nonlinear partial differential equations},
  author    = {Raissi, Maziar and Perdikaris, Paris and Karniadakis, George Em},
  journal   = {Journal of Computational Physics},
  volume    = {378},
  pages     = {686--707},
  year      = {2019},
  publisher = {Elsevier}
}

@article{karniadakis2021physics,
  title     = {Physics-informed machine learning},
  author    = {Karniadakis, George Em and Kevrekidis, Ioannis G. and Lu, Lu and Perdikaris, Paris and Wang, Sifan and Yang, Liu},
  journal   = {Nature Reviews Physics},
  volume    = {3},
  number    = {6},
  pages     = {422--440},
  year      = {2021},
  publisher = {Nature Publishing Group}
}

@article{li2025robotic,
  title   = {Robotic world model: A neural network simulator for robust policy optimization in robotics},
  author  = {Li, Chenhao and Krause, Andreas and Hutter, Marco},
  journal = {arXiv preprint arXiv:2501.10100},
  year    = {2025}
}

@inproceedings{hinterstoisser2012model,
  title     = {Model Based Training, Detection and Pose Estimation of Texture-Less {3D} Objects in Heavily Cluttered Scenes},
  author    = {Hinterstoisser, Stefan and Lepetit, Vincent and Ilic, Slobodan and Holzer, Stefan and Bradski, Gary and Konolige, Kurt and Navab, Nassir},
  booktitle = {Asian Conference on Computer Vision (ACCV)},
  pages     = {548--562},
  year      = {2012},
  publisher = {Springer}
}

@article{mittal2023orbit,
  author  = {Mittal, Mayank and Yu, Calvin and Yu, Qinxi and Liu, Jingzhou and Rudin, Nikita and Hoeller, David and Yuan, Jia Lin and Singh, Ritvik and Guo, Yunrong and Mazhar, Hammad and Mandlekar, Ajay and Babich, Buck and State, Gavriel and Hutter, Marco and Garg, Animesh},
  title   = {{Orbit}: A Unified Simulation Framework for Interactive Robot Learning Environments},
  journal = {IEEE Robotics and Automation Letters},
  volume  = {8},
  number  = {6},
  pages   = {3740--3747},
  year    = {2023},
  doi     = {10.1109/LRA.2023.3270034}
}

@misc{nvidia_isaacsim,
  author       = {{NVIDIA Corporation}},
  title        = {{NVIDIA Isaac Sim}},
  howpublished = {\url{https://developer.nvidia.com/isaac/sim}},
  year         = {2024},
  note         = {Accessed: 2026-11-07}
}

@article{haddadin2022franka,
  author  = {Haddadin, Sami and Parusel, Sven and Johannsmeier, Lars and Golz, Saskia and Gabl, Simon and Walch, Florian and Sabaghian, Mohamadreza and J{\"a}hne, Christoph and Hausperger, Lukas and Haddadin, Simon},
  title   = {The {Franka Emika} Robot: A Reference Platform for Robotics Research and Education},
  journal = {IEEE Robotics \& Automation Magazine},
  volume  = {29},
  number  = {2},
  pages   = {46--64},
  year    = {2022},
  doi     = {10.1109/MRA.2021.3138382}
}

@book{greenwood2006advanced,
  title     = {Advanced Dynamics},
  author    = {Greenwood, Donald T.},
  year      = {2003},
  doi       = {10.1017/CBO9780511800207},
  publisher = {Cambridge University Press}
}

@book{de2012theory,
  title     = {Theory of Robot Control},
  editor    = {Canudas de Wit, Carlos and Siciliano, Bruno and Bastin, Georges},
  year      = {1996},
  publisher = {Springer},
  address   = {London},
  doi       = {10.1007/978-1-4471-1501-4}
}

@book{featherstone2007rigid,
  title     = {Rigid Body Dynamics Algorithms},
  author    = {Featherstone, Roy},
  year      = {2008},
  publisher = {Springer},
  address   = {Boston, MA},
  doi       = {10.1007/978-1-4899-7560-7}
}

\appendix
\section{Model Architecture}
\label{sec:app-model}

\subsection{Context Encoder}
\label{sec:app-encoder}

\paragraph{Input and architecture.}
The encoder receives the most recent \(K=12\) control steps
(\(240\,\mathrm{ms}\)) of proprioception. Each frame contains the seven
arm-joint positions \(q_k\), measured relative to the default Franka pose,
the joint velocities \(\dot q_k\), and the torque \(\bar\tau_{k-1}\) applied
over the preceding interval, where \(\bar\tau_k\) denotes the
interval-averaged torque \(\tau_k\) of Section~\ref{sec:problem}. Thus, each
recorded state is paired with the torque that produced it:
\begin{equation}
    u_t
    =
    \operatorname{vec}
    \bigl[(q_k,\dot q_k,\bar\tau_{k-1})\bigr]_{k=t-K+1}^{t}
    \in\mathbb{R}^{K\cdot 21}
    =
    \mathbb{R}^{252}.
    \label{eq:app-encoder-input}
\end{equation}
The torque \(\bar\tau_t\) applied during the first prediction interval is not
available to the encoder; it enters the model only through the forward dynamics (Section~\ref{sec:conditioning}). The input frames are flattened in temporal order and
provided in physical units (rad, rad/s, and N\,m), without normalization. The
encoder does not receive the cube pose, grasp offset, payload mass, or CoM.

The encoder \(E_\phi\) is an MLP with two hidden layers of width 256 and SiLU
activations. Its two output heads produce the context mean and log variance:
\begin{align}
    h &=
    \operatorname{SiLU}\!\left(
        W_2\operatorname{SiLU}(W_1u_t+b_1)+b_2
    \right),\\
    \mu_t &=
    10\tanh\!\left(\frac{W_\mu h+b_\mu}{10}\right),\\
    \log\sigma_t^2 &=
    \operatorname{clip}(W_\sigma h+b_\sigma,-10,5),
\end{align}
where \(W_1\in\mathbb{R}^{256\times252}\),
\(W_2\in\mathbb{R}^{256\times256}\), and
\(W_\mu,W_\sigma\in\mathbb{R}^{d_z\times256}\), with \(d_z=4\).
The soft bound on \(\mu_t\) and the clipping of \(\log\sigma_t^2\) keep the
context finite; the same bound is applied again to the sampled context below. The encoder has 132,616 parameters. Its linear layers use PyTorch's default
initialization, and it has no normalization or dropout layers.

\paragraph{Context sampling and use.}
During training, one context is sampled per window and held fixed over the
training rollout of 8 steps:
\begin{equation}
    z_t =
    10\tanh\!\left(
        \frac{\mu_t+\sigma_t\odot\epsilon}{10}
    \right),
    \qquad \epsilon\sim\mathcal{N}(0,I).
\end{equation}
The Gaussian distribution is regularized toward \(\mathcal{N}(0,I)\) using a
KL term with weight \(\lambda_{\mathrm{KL}}=10^{-4}\). The contrastive term is
applied to the encoder output. The prediction loss also backpropagates through
\(z_t\), so the encoder is trained end-to-end with the dynamics.

At evaluation, we use \(z_t=\mu_t\). The context is recomputed from the
sliding history at the start of each prediction and held fixed throughout the
\(N=15\)-step rollout.
\subsection{FiLM-Conditioned Lagrangian Networks}
\label{sec:app-film}

\paragraph{Network architecture.}
Three separate networks produce the diagonal entries of \(A\), the
strictly lower-triangular entries of \(A\), and the potential \(V\). Each
network takes the absolute joint configuration \(q\in\mathbb{R}^7\) as input
and has three hidden layers of width 1024, the last two of which are
modulated. None of the networks takes joint
velocity as input. For each network, the forward pass is
\begin{align}
    h_1 &=
    \operatorname{SiLU}(W_0q+b_0),\\
    a_\ell &=
    \bigl(1+\psi(\gamma_\ell(z))\bigr)
    \odot(W_\ell h_\ell+b_\ell)
    +\psi(\beta_\ell(z)),
    \qquad \ell=1,2,\\
    h_{\ell+1} &= \operatorname{SiLU}(a_\ell),
    \qquad \ell=1,2,\\
    y &= W_{\mathrm{out}}h_3+b_{\mathrm{out}}.
\end{align}
This is the gated residual form of the FiLM update in
Section~\ref{sec:conditioning}: the scale and shift enter as
\(1+\psi(\gamma_\ell(z))\) and \(\psi(\beta_\ell(z))\), so that zero maps leave the
backbone unchanged.
The output dimension is 7 for the diagonal of \(A\), 21 for its
strictly lower-triangular part, and 1 for \(V\). The diagonal is constrained
to be positive:
\begin{equation}
    A_{ii}=\operatorname{softplus}(y_i)+10^{-3},
    \qquad
    H=AA^\top+10^{-4}I,
\end{equation}
where the small diagonal term is added to the construction of
Section~\ref{sec:preliminaries} for numerical stability.
The input and output layers are not modulated. Each network has about
\(2.13\) million parameters, or \(6.41\) million across all three networks.

\paragraph{FiLM modulation.}
At each modulated layer, the scale and shift are affine functions of the context:
\begin{equation}
    \gamma_\ell(z)=\Gamma^\gamma_\ell z+\gamma^0_\ell,
    \qquad
    \beta_\ell(z)=\Gamma^\beta_\ell z+\beta^0_\ell,
\end{equation}
where each map is from \(\mathbb{R}^4\) to \(\mathbb{R}^{1024}\). Each network
and layer has its own maps, for 61,440 parameters in total, approximately
\(1\%\) of the Lagrangian-network parameters.

Both maps use the same bounded gate:
\begin{equation}
    \psi(u)=\alpha T\tanh(u/T),
    \qquad
    \alpha=0.3,\quad T=3.
\end{equation}
The gate has slope \(\alpha\) at the origin and saturates at
\(\pm\alpha T=\pm0.9\). Consequently, each hidden-unit gain lies in
\((0.1,1.9)\), and each bias shift lies in \((-0.9,0.9)\). The gain cannot
switch a unit off or change its sign. Increasing \(T\) moves the saturation
point outward without changing the slope at the origin, allowing the context
to produce larger feature changes for payloads near or beyond the training
range before saturation.

All FiLM maps are initialized to zero. Thus, at initialization, each
conditioned network is identical to the same unconditioned DeLaN for every
\(z\). The encoder receives a gradient through the FiLM maps once the context
becomes useful for prediction. All other weights use PyTorch's default
initialization.

\paragraph{Configuration derivatives.}
The Coriolis and gravity terms require
\(\partial A/\partial q\) and \(\partial V/\partial q\) at fixed \(z\).
Because \(z\) remains constant within a window, the FiLM modulation is
constant with respect to \(q\). We therefore propagate the configuration
Jacobian alongside the forward pass:
\begin{align}
    \frac{\partial h_{\ell+1}}{\partial q}
    &=
    \operatorname{diag}\!\bigl(\operatorname{SiLU}'(a_\ell)\bigr)
    \operatorname{diag}\!\bigl(1+\psi(\gamma_\ell(z))\bigr)
    W_\ell
    \frac{\partial h_\ell}{\partial q},\\
    \frac{\partial h_1}{\partial q}
    &=
    \operatorname{diag}\!\bigl(
        \operatorname{SiLU}'(W_0q+b_0)
    \bigr)W_0.
\end{align}
The shift \(\beta_\ell\) affects these derivatives through
\(\operatorname{SiLU}'(a_\ell)\). The resulting Jacobians give
\(\partial H/\partial q\) and \(g=\partial V/\partial q\) in a single
forward pass, without nested automatic differentiation. Gradients with
respect to the network weights, FiLM maps, and encoder are then computed by
backpropagation through the full construction.

\section{Baselines}
\label{sec:baselines}

\subsection{Common protocol}

All methods use the same 318 training episodes and windows: $K = 12$ history
frames of $(q_k, \dot q_k, \bar\tau_{k-1})$ ($q_k$ relative to the default
Franka pose), followed by the recorded future torques. No method receives the
cube pose, grasp offset, a nominal arm model or payload properties. All are
trained for 40 epochs with an 8-step teacher-forced horizon on batches of 256
(512 for DeLaN), and the final weights are evaluated by a 15-step (300\,ms)
open-loop rollout under the recorded torques, with any inferred context held
fixed. The cube pose follows from forward kinematics with the grasp transform
at the last history frame.

Acceleration-output models are integrated with the trapezoidal update of
Appendix~\ref{sec:app-discrete-time-dynamics} and trained on
$\bar{\ddot q}_k = (\dot q_{k+1} - \dot q_k)/\Delta t$ (standardized per joint).
RWM and CaDM keep their released next-state output. Since one 20\,ms step
cannot reproduce the simulator's 40 substeps even with the oracle acceleration
(Appendix~\ref{sec:app-discrete-time-dynamics}), this choice, if anything,
disadvantages our model. Otherwise each baseline keeps its released network
and optimizer (Table~\ref{tab:baselines}); planning, policy and
data-collection components are not used.

\begin{table}[h]
\centering
\small
\begin{tabular}{lllrl}
\toprule
Method & Context from history & Output & Parameters & Optimizer \\
\midrule
PICR       & MLP encoder, $z \in \mathbb R^4$, FiLM & $\bar{\ddot q}$ (Lagrangian) & 6,545,957 & Adam $3{\times}10^{-4}$, cosine \\
RWM        & GRU state, 256-d                  & next state & 956,184 & Adam $10^{-3}$, clip 1 \\
CaDM       & MLP encoder, $e \in \mathbb R^{10}$ & state change & 231,008$^\dagger$ & Adam $10^{-3}$ \\
CaDeLaC    & LSTM encoder, $z \in \mathbb R^{10}$, concat & $\bar{\ddot q}$ (Lagrangian) & 6,719 & Adam $10^{-3}$, clip 1 \\
DeLaN      & none                              & $\bar{\ddot q}$ (Lagrangian) & 12,125 & Adam $10^{-4}$ \\
Neural-Fly & least-squares weights $a \in \mathbb R^{3\times7}$ & $\bar{\ddot q}$ & 133,719$^\dagger$ & Adam $5{\times}10^{-4}$ \\
\bottomrule
\end{tabular}
\caption{Baselines. Parameters count the prediction path. $^\dagger$CaDM's
backward core (129,814) and Neural-Fly's discriminator (248) are used only in
training.}
\label{tab:baselines}
\end{table}

\subsection{RWM}

RWM~\citep{li2025robotic}: a two-layer GRU reads $[q_k, \dot q_k, \bar\tau_k]$
and five Gaussian heads ($256 \to 128 \to 14$) predict the next state
residually; the prediction is the ensemble mean, fed back with the next
recorded torque. We use the released bootstrapped objective with its log-std
bound term.

\subsection{CaDM}

CaDM~\citep{lee2020context}, reimplemented in PyTorch from the TensorFlow~1
release. An MLP ($256$--$128$--$64$) encodes the 11 history transitions
$(x_{j+1} - x_j, \bar\tau_j)$, $x = (q, \dot q)$. Forward and backward cores
(four 200-unit swish layers) predict the normalized state change, with the
released loss (forward $+\,0.5\,\times$ backward MSE, plus L2). We use the
deterministic single-model variant.

\subsection{CaDeLaC}

CaDeLaC~\citep{schulze2025context}: a five-layer, 10-unit LSTM encodes the
history, and one MLP ($30$--$20$, ELU) on $[q, z]$ outputs the Cholesky factor
of $H$ and $V$, with the release's sizes. Instead of the release's residual
over a nominal Pinocchio model, it models the full system from raw torques.

\subsection{DeLaN}

DeLaN~\citep{lutter2019deep}, following the JAX release: two MLPs (two 64-unit
tanh layers) on $[\cos q, \sin q]$ output the Cholesky factor of $H$ and $V$.

\subsection{Neural-Fly}

Neural-Fly~\citep{o2022neural} predicts
$\hat y_k = f_0(\xi_k) + \phi(\xi_k)^\top a$ with the released basis $\phi$ and
a learned nominal $f_0$ (four 200-unit swish layers) in place of the release's
physics model. $a$ is solved by ridge least squares ($\lambda = 10^{-3}$) on
the 11 history transitions. Training uses the released domain-adversarial
meta-learning, with a discriminator on the payload-mass level (training only,
as for PICR's contrastive term).

\section{Ablation}
\label{sec:app-ablation}
\subsection{Effect of the contrastive objective}
\label{sec:app-ablation-con}

The contrastive term of Section~\ref{sec:contrastive} is part of the PICR
training objective. To measure its contribution, we train a variant with
$\lambda_{\mathrm{con}} = 0$ and leave everything else unchanged: the
training split, the context encoder and its KL regularization, the
FiLM-conditioned Lagrangian networks, the optimizer and the 40-epoch
schedule. Without the term, the two-view episode batches are not needed,
and the variant uses standard shuffled batches with the same number of
batches per epoch. The variant therefore differs from PICR in whether
the payload labels shape the context during training. Both models are
trained with seed 0 and evaluated at the final epoch.

We use two evaluations. Prediction accuracy is the $N=15$ cube ADD on
the figure-eight payload sweep of Appendix~\ref{sec:app-payload-sweep},
split into the training mass range (0.1--2\,kg, 16,071 windows) and the
unseen range (2--4\,kg, 16,896 windows). Component accuracy is the RMSE
of the joint-2 gravity, inertial and Coriolis terms against the simulator
references of Section~\ref{sec:dynamics-term-reconstruction}, over the
first second of a fixed-payload carry with a 3\,kg payload, outside the
training range.

\begin{table}[h]
\centering
\small
\begin{tabular}{lccc}
\toprule
 & \multicolumn{2}{c}{Figure-eight ADD [cm] $\downarrow$} & \\
\cmidrule(lr){2-3}
 & 0.1--2\,kg & 2--4\,kg
 & $g_2$ / $[H\ddot q]_2$ / $c_2$ RMSE [N\,m] $\downarrow$ \\
\midrule
PICR                 & \textbf{4.58} & \textbf{6.92}
                     & \textbf{1.84} / \textbf{1.63} / 1.46 \\
PICR w/o contrastive & 6.07 & 9.78
                     & 2.11 / 2.87 / \textbf{1.15} \\
\bottomrule
\end{tabular}
\caption{Removing the contrastive term from PICR
($\lambda_{\mathrm{con}} = 0$; seed 0, final epoch). Term RMSE is for
joint 2 on a 3\,kg carry.}
\label{tab:ablation-con}
\end{table}

Removing the contrastive term increases the ADD in both mass ranges, from
4.58 to 6.07\,cm within the training range and from 6.92 to 9.78\,cm
beyond it (Table~\ref{tab:ablation-con}). It also increases the error of
the two mass-dependent components on the 3\,kg carry, from 1.84 to
2.11\,N\,m for gravity and from 1.63 to 2.87\,N\,m for the inertial
term. Without the label-based organization, the context is shaped only by
the dynamics losses and is less effective at encoding the payload
condition that enters $H$ and $V$. The Coriolis term is the exception:
the variant without the contrastive term attains a lower error (1.15
versus 1.46\,N\,m). This term is small, depends on the payload only
through the derivatives of $H$, and the comparison uses a single seed.
We therefore interpret the contrastive term as improving prediction
accuracy and the recovery of the mass-dependent components, rather than
as uniformly improving all learned terms.

\subsection{Additional supervision to baselines}
\label{sec:app-supcon-baselines}

The contrastive term of PICR defines positives from the simulator's payload
mass and center of mass. These are training-only labels that the baselines do
not receive. To test whether this extra information alone explains the gap,
we added the identical term to every baseline that has a context
representation, using the same weight ($\lambda_{\mathrm{con}} = 0.1$), the
same temperature ($\kappa = 0.1$), the same Gaussian positive weighting
($\sigma_m = 0.271$\,kg, $\sigma_r = 1$\,cm) and the same two-view episode
batches. The term acts on CaDeLaC's 10-dimensional latent, on CaDM's
10-dimensional context, and on RWM's 256-dimensional GRU output after the
12-frame history, with no projection head. DeLaN has no context and is
excluded. All other settings, including data, objectives, optimizers and the
40-epoch schedule, are unchanged. Each row is a single seed evaluated at the
final epoch.

We evaluate on 3D figure-eight tracking with a constant payload. The PPO
policy follows the figure-eight of Appendix~\ref{sec:app-payload-sweep} while the lap
period shortens linearly from 3\,s to 1.5\,s over 12\,s, which produces
faster and more torque-intensive motion than the carry episodes. Payload
masses are either within the training range (0.1--2.0\,kg, 125,865 windows)
or beyond it (2.0--4.0\,kg, 61,522 windows). We also report the torque-intervention gain.

\begin{table}[h]
\centering
\small
\begin{tabular}{lccc}
\toprule
 & \multicolumn{2}{c}{Figure-eight ADD [cm]} & \\
\cmidrule(lr){2-3}
 & 0.1--2\,kg & 2--4\,kg & Gain at 20\,N\,m \\
\midrule
RWM                   & 10.58 & 13.57 & 0.73 \\
RWM + contrastive     & 12.48 & 13.21 & 0.55 \\
\midrule
CaDM                  & 9.76 & 10.56 & 0.63 \\
CaDM + contrastive    & 11.06 & 11.66 & 0.67 \\
\midrule
CaDeLaC               & 11.33 & 22.16 & 0.16 \\
CaDeLaC + contrastive & 10.48 & 17.94 & 0.22 \\
\midrule
PICR                  & 7.21 & 6.30 & 0.86 \\
\bottomrule
\end{tabular}
\caption{Adding PICR's contrastive term to the baselines (seed 0, $N = 15$).
The gain is the median one-step projected acceleration gain at $|\Delta\tau| = 20$\,N\,m, averaged over
joints 1--3.}
\label{tab:supcon-baselines}
\end{table}

The contrastive term did not consistently help the baselines
(Table~\ref{tab:supcon-baselines}). It degraded RWM within the training
range (ADD 10.58 to 12.48\,cm) and changed it little beyond it (13.57 to
13.21\,cm), while its gain fell from 0.73 to 0.55. It also degraded CaDM
(9.76 to 11.06\,cm and 10.56 to 11.66\,cm), although its gain rose slightly
(0.63 to 0.67). CaDeLaC improved (11.33 to 10.48\,cm and 22.16 to
17.94\,cm; gain 0.16 to 0.22). With the labels, every
baseline still remains behind PICR in both mass ranges and in gain. The
labels therefore do not, on their own, account for PICR's advantage.
In PICR, the representation shaped by the contrastive term is the context
$z$ that modulates the Lagrangian networks through FiLM, so the
organization of $z$ by payload enters the inertia, Coriolis and gravity
terms directly. In the baselines, the same term shapes a representation that
conditions a predictor without physical structure (RWM, CaDM) or through
input concatenation (CaDeLaC). There it can act as an additional
objective that competes with the prediction objective, as it did for RWM and
CaDM. CaDeLaC, whose context also enters a Lagrangian network, benefits from
the term, but its error beyond the training range remains almost three times
that of PICR (17.94 versus 6.30\,cm).

\subsection{Effect of the conditioning mechanism}
\label{sec:app-ablation-cond}

PICR conditions the Lagrangian networks on the context through FiLM. To
test whether this choice matters, we replace FiLM with three alternative
ways for $z$ to determine $H$ and $V$, keeping everything else fixed: the
context encoder with its KL and contrastive terms, the three networks
for the diagonal and strictly lower-triangular entries of $A$ and for
$V$ (three hidden layers of width 1024, input $q$), the construction
$H = AA^\top + 10^{-4}I$, the derivation of $c$ from $H$ and of $g$ from
$\nabla_q V$, the losses $\mathcal L_{\ddot q}$, $\mathcal L_{\mathrm{ID}}$
and $\mathcal L_{\mathrm{con}}$, the data and the 40-epoch schedule.
Recall that FiLM modulates each modulated layer $\ell = 1, 2$ as
$a_\ell = (1 + \psi(\gamma_\ell(z))) \odot (W_\ell h_\ell + b_\ell)
+ \psi(\beta_\ell(z))$. The alternatives are:
\begin{itemize}
    \item \textbf{Input concatenation.} The context enters only the
    first layer, $h_1 = \operatorname{SiLU}(W_0 [q; z] + b_0)$, and the
    hidden layers are unmodulated.
    \item \textbf{Shift only.} Each modulated layer receives an additive
    term, $a_\ell = W_\ell h_\ell + b_\ell + \psi(U_\ell z)$. This is
    approximately equivalent to concatenating $z$ to the input of every hidden layer,
    and corresponds to FiLM without its multiplicative scale.
    \item \textbf{Low-rank hypernetwork.} The context generates a rank-4
    update of each hidden weight matrix,
    $W_\ell(z) = W_\ell + P_\ell \operatorname{diag}(\psi(R_\ell z))
    Q_\ell^\top$. Unlike FiLM, which rescales individual units, this
    changes how units are mixed.
\end{itemize}
All context pathways are initialized to zero, so that every variant
starts from the same unconditioned DeLaN as PICR. The bounded gate $\psi$
is the one used by FiLM (Appendix~\ref{sec:app-film}). The added
parameters are about 12k, 31k and 49k for the three variants, compared
with 61k for FiLM. Each variant is trained with three seeds. We report
the figure-eight ADD and the joint-2 term RMSE on the 3\,kg carry, as in
Appendix~\ref{sec:app-ablation-con}.

\begin{table}[h]
\centering
\small
\begin{tabular}{lccc}
\toprule
 & \multicolumn{2}{c}{Figure-eight ADD [cm] $\downarrow$} & \\
\cmidrule(lr){2-3}
 & 0.1--2\,kg & 2--4\,kg
 & $g_2$ / $[H\ddot q]_2$ / $c_2$ RMSE [N\,m] $\downarrow$ \\
\midrule
FiLM (PICR)       & $4.93 \pm 0.54$ & $\mathbf{6.76 \pm 1.28}$
                  & \textbf{2.15} / 2.31 / \textbf{1.59} \\
Input concat      & $4.45 \pm 0.37$ & $8.49 \pm 0.69$
                  & 4.02 / 2.27 / 2.09 \\
Shift only        & $4.59 \pm 0.54$ & $8.43 \pm 0.80$
                  & 3.33 / 2.29 / 1.96 \\
Low-rank hypernet & $\mathbf{4.26 \pm 0.35}$ & $11.75 \pm 2.63$
                  & 5.85 / \textbf{1.91} / 3.02 \\
\bottomrule
\end{tabular}
\caption{Conditioning mechanisms for the Lagrangian networks. ADD is
the mean $\pm$ standard deviation over three seeds; term RMSE is the
seed mean. Final epoch. The FiLM row corresponds to PICR in Table~\ref{tab:causal-results}; its seed 0 is the model of Tables~\ref{tab:ablation-con} and~\ref{tab:ablation-capacity}.}
\label{tab:ablation-cond}
\end{table}

Table~\ref{tab:ablation-cond} shows that FiLM generalizes best beyond the training range. On unseen payloads (2--4\,kg), FiLM attains an ADD of 6.76\,cm, whereas the alternatives reach 8.4--11.8\,cm, an increase of 25--74\%. Within the training range, all four mechanisms lie within the variation across seeds (4.3--4.9\,cm), so the gain from FiLM comes precisely where adaptation is needed. The component errors identify its source. On the unseen 3\,kg carry, FiLM reduces the gravity error from 3.3--5.9 to 2.15\,N\,m and the Coriolis error from 2.0--3.0 to 1.59\,N\,m. The context therefore reshapes the potential $V$, and hence $g$, in a way that extrapolates with payload mass.

The shift-only variant isolates the role of the multiplicative scale. Removing it raises the unseen ADD from 6.76 to 8.43\,cm and the gravity error from 2.15 to 3.33\,N\,m, indicating that multiplicative modulation carries the mass dependence beyond the training range. Greater expressiveness does not substitute for it: the low-rank hypernetwork, which alters how units are mixed rather than rescaling them, yields the highest unseen ADD (11.75\,cm) and gravity error (5.85\,N\,m).

\subsection{Effect of model capacity}
\label{sec:app-ablation-capacity}

The baselines keep the network sizes of their releases
(Table~\ref{tab:baselines}), so the models compared in
Table~\ref{tab:causal-results} range from 6.7k parameters (CaDeLaC) to
6.5M (PICR). To check that the comparison does not rest on this
difference, we bring the three Lagrangian models to a common size of
about 0.5M parameters by changing layer widths only: the three
Lagrangian networks of PICR from 1024 to 256 units (556k parameters),
the core of CaDeLaC from (30, 20) to (640, 640) units (445k), and the
two networks of DeLaN from 64 to 512 units per layer (556k). RWM
(0.96M) and CaDM (0.23M) already lie in this range and are unchanged.
Encoders, losses, optimizers, batch sizes and the 40-epoch schedule are
those of the respective rows; the wider DeLaN uses the gradient clip of
CaDeLaC (1.0), without which it diverged in preliminary runs. Accuracy
is the $N=15$ cube ADD on the figure-eight payload sweep, split as in
Table~\ref{tab:ablation-con}. The narrow PICR is trained with three
seeds, the widened baselines with one.

\begin{table}[h]
\centering
\small
\begin{tabular}{lrcc}
\toprule
 & & \multicolumn{2}{c}{Figure-eight ADD [cm] $\downarrow$} \\
\cmidrule(lr){3-4}
 & Parameters & 0.1--2\,kg & 2--4\,kg \\
\midrule
PICR                 & 6.55M & \textbf{4.58}          & \textbf{6.92} \\
PICR, width 256      & 0.56M & $4.77 \pm 0.19$        & $8.66 \pm 1.52$ \\
\midrule
RWM                  & 0.96M & 7.09                   & 11.63 \\
CaDM                 & 0.23M & 6.79                   & 10.90 \\
\midrule
CaDeLaC              & 6.7k  & 9.07                   & 24.75 \\
CaDeLaC, width 640   & 0.45M & 7.49                   & 16.76 \\
\midrule
DeLaN                & 12k   & 6.82                   & 25.65 \\
DeLaN, width 512     & 0.56M & 7.22                   & 23.70 \\
\bottomrule
\end{tabular}
\caption{Lagrangian models at a common size of about 0.5M parameters.
Only layer widths differ from the rows of Table~\ref{tab:causal-results} (seed 0);
the narrow PICR is the mean $\pm$ standard deviation over three seeds.
Final epoch.}
\label{tab:ablation-capacity}
\end{table}

Narrowing PICR twelvefold raises the ADD from 4.58 to 4.77\,cm within the
training range and from 6.92 to 8.66\,cm beyond it, and it remains the
most accurate model in both ranges (Table~\ref{tab:ablation-capacity}).
Widening CaDeLaC improves it in both ranges (9.07 to 7.49\,cm and 24.75
to 16.76\,cm) but leaves it 2.7 and 8.1\,cm behind the narrow PICR;
widening DeLaN changes little (6.82 to 7.22\,cm and 25.65 to 23.70\,cm).
The ranking of Table~\ref{tab:causal-results} is therefore not a
consequence of model size, and the gap on unseen payloads persists when
the Lagrangian models have the same number of parameters.

\section{Data Collection}
\label{sec:app-impl}

\subsection{Simulation and Data Recording}
\label{sec:app-data-recording}

All data are recorded with a Franka Panda in Isaac Lab~2.3.2 (Isaac Sim~5.1,
PhysX) with a \(0.5\,\mathrm{ms}\) physics step and a
\(\Delta t=20\,\mathrm{ms}\) control interval (40 substeps). Each step stores
\(q,\dot q\) of the seven arm joints and the substep-averaged measured torque
\(\bar\tau=\tfrac{1}{40}\sum_{i=1}^{40}\tau_i^{\mathrm{PhysX}}\), which includes
the actuator command, implicit-drive forces and velocity-limit constraint
forces; unlike the commanded effort, it is consistent with the recorded
transition.

The payload is a cube of side \(s=4.8\,\mathrm{cm}\), mass \(m\), CoM \(r\) and
inertia \(\tfrac16 ms^2 I_3\). Its pose is recorded for evaluation only. For
the evaluation sets we also record the arm's \(H_{\mathrm{arm}}\ddot q\),
\(c_{\mathrm{arm}}\) and \(g_{\mathrm{arm}}\) from PhysX, averaged per step and
combined with analytic payload terms to form the references of
Section~\ref{sec:dynamics-term-reconstruction}.

All sets keep only carry-phase steps (cube lifted and held), dropping the first
three steps of each carry segment. An episode is removed if the cube moves
more than \(10\,\mathrm{mm}\) from its median in-hand position within any
20-step window.

\label{sec:app-dataset-overview}
\begin{table}[t]
    \centering
    \small
    \begin{tabular}{
        @{}
        >{\raggedright\arraybackslash}p{0.14\linewidth}
        *{3}{>{\raggedright\arraybackslash}p{\dimexpr(0.86\linewidth-6\tabcolsep)/3\relax}}
        @{}
    }
        \toprule
        Set & Motion and control & Payload condition & Size and use \\
        \midrule
        Training
        & Joint excitation; PID with OU torque perturbation
        & \(0.1\)--\(2.0\,\mathrm{kg}\), 8 levels; \(5^3\) CoM lattice
        & 318 training / 40 validation episodes \\
        Payload sweep
        & PPO figure-eight tracking
        & Mass ramp from \(0.1\) to \(4.0\,\mathrm{kg}\)
        & 33 episodes / 32,967 windows \\
        Torque intervention
        & PPO replay with torque offsets
        & \(0.1\)--\(2.0\,\mathrm{kg}\); grasp offset along hand \(x\)
        & 1,887 of 5,616 pairs \\
        \bottomrule
    \end{tabular}
    \caption{Datasets used in this work. Only the training split is used to
    fit the models. Each window contains a 12-step history and a 15-step
    prediction horizon.}
    \label{tab:datasets}
\end{table}

\subsection{Training Data: Joint Excitation}
\label{sec:app-excitation}

The training data comprise 400 episodes (four batches of 100 parallel
environments), 50 per mass
\(m \in \{0.10,0.37,0.64,0.91,1.19,1.46,1.73,2.00\}\,\mathrm{kg}\), with the CoM
sampled as in Section~\ref{sec:app-com-grasp}. The arm is torque-controlled;
the fingers are held closed by implicit PD drives. Each 23\,s episode consists
of a scripted grasp and lift, a 5\,s quintic move to a start pose \(q_0\), a
0.5\,s settle, and a joint-wise cosine sweep
\begin{equation}
    q^{\mathrm{ref}}_j(t) = q_{0,j}+s_j A_j\bigl(1-\cos\omega_j t\bigr),
    \qquad j=1,\ldots,7,
    \label{eq:app-cosine-sweep}
\end{equation}
which moves joint \(j\) by \(2A_j\) toward its limit in direction
\(s_j\in\{\pm1\}\).

\paragraph{Trajectory generation.}
Sweeps are generated offline with Pinocchio and hpp-fcl (Franka collision
meshes plus the cube). \(q_0\) is sampled from the joint box shrunk by
\(0.12\,\mathrm{rad}\) and must be collision-free with \(5\,\mathrm{cm}\)
clearance and reachable by a straight collision-free path. Per joint,
\(s_j\) is random, \(2A_j = u_j d_j\) with \(u_j\sim\mathcal U(0.05,1)\) and
\(d_j\) the distance to the limit, and the target peak velocity is
\(v_j=(0.97-0.82u^2)\dot q_{\max,j}\), \(u\sim\mathcal U(0,1)\) (median
\(0.77\,\dot q_{\max,j}\)), enlarging \(A_j\) if needed so that
\(A_j\geq v_j^2/\ddot q_{\max,j}\). Then \(\omega_j=v_j/A_j\), capped by
\(\sqrt{\ddot q_{\max,j}/A_j}\) and \(12\,\mathrm{rad/s}\), floored at one
period per 12\,s, and rounded to a multiple of \(2\pi/60\,\mathrm{s^{-1}}\).
Velocity limits are \(2.175\) / \(2.61\,\mathrm{rad/s}\) and acceleration
limits \(12\) / \(18\,\mathrm{rad/s^2}\) for joints 1--4 / 5--7.

Trajectories violating joint limits, TCP height \(\ge 0.20\,\mathrm{m}\),
joint 3--7 height \(\ge 0.12\,\mathrm{m}\), horizontal TCP distance
\(\ge 0.25\,\mathrm{m}\), TCP acceleration \(\le 8\,\mathrm{m/s^2}\), or
self-collision have all amplitudes scaled down by bisection. Of ten
candidates per start pose we keep the one maximizing
\(\sum_j \max_t |\dot q_j|/\dot q_{\max,j}\). The episodes use trajectories
0--399 of a 1,600-trajectory pool.

\paragraph{Joint control and torque perturbation.}
At every substep, a PID controller with arm-only computed-torque
feed-forward tracks the reference:
\begin{equation}
    \tau = K_p e + K_i\!\int\! e\,dt + K_d(\dot q^{\mathrm{ref}}-\dot q)
    + \hat H\ddot q^{\mathrm{ref}} + \hat c + \hat g + \eta_k,
    \qquad e=q^{\mathrm{ref}}-q,
    \label{eq:app-pid}
\end{equation}
with \(K_p=\operatorname{diag}(600,600,600,600,250,150,50)\), \(K_i=3K_p\),
\(K_d=\operatorname{diag}(20,50,20,30,8,8,1)\). The payload is not
compensated. The integral term is limited to \(60\) / \(5\,\mathrm{N\,m}\)
with anti-windup and the command clipped to \(87\) / \(12\,\mathrm{N\,m}\)
(joints 1--4 / 5--7). To decorrelate torque from the state history, we add an
Ornstein--Uhlenbeck perturbation, updated once per control step:
\begin{equation}
    \eta_{k+1} = a\eta_k+\sqrt{1-a^2}\,\sigma\odot\epsilon_k,
    \quad \epsilon_k\sim\mathcal N(0,I),
    \quad a=e^{-\Delta t/\tau_c},
    \label{eq:app-ou}
\end{equation}
with \(\tau_c=0.2\,\mathrm{s}\) and
\(\sigma=(8,10,8,10,1.6,2.4,0.07)\,\mathrm{N\,m}\), from the start of the
joint-space motion. The command saturates in \(1.6\%\) of steps (mostly
joint 2), \(0.5\%\) reach a velocity limit, and the cube moves at most
\(3.3\,\mathrm{mm}\) in hand.

\paragraph{Episode selection and coverage.}
358 of 400 episodes keep the cube above \(0.15\,\mathrm{m}\) for at least 60
steps (none fail the slip filter), giving 364,057 carry steps. Five episodes
per mass are held out for validation (39,908 windows); the remaining 318
(323,389 steps) form the training split. These steps cover \(98.3\%\) of the
per-joint \((q,\dot q)\) bins
(\(0.1\,\mathrm{rad}\times0.2\,\mathrm{rad/s}\)) reachable by a cosine sweep
(95--99\% per joint) and \(98.6\%\) of collision-free pairwise position bins
(\(0.2\,\mathrm{rad}\); lowest 92.9\% for joints 2--3).

\subsection{Center-of-Mass and Grasp-Offset Variation}
\label{sec:app-com-grasp}

\begin{figure}[t]
  \centering
  \includegraphics[width=1\linewidth]{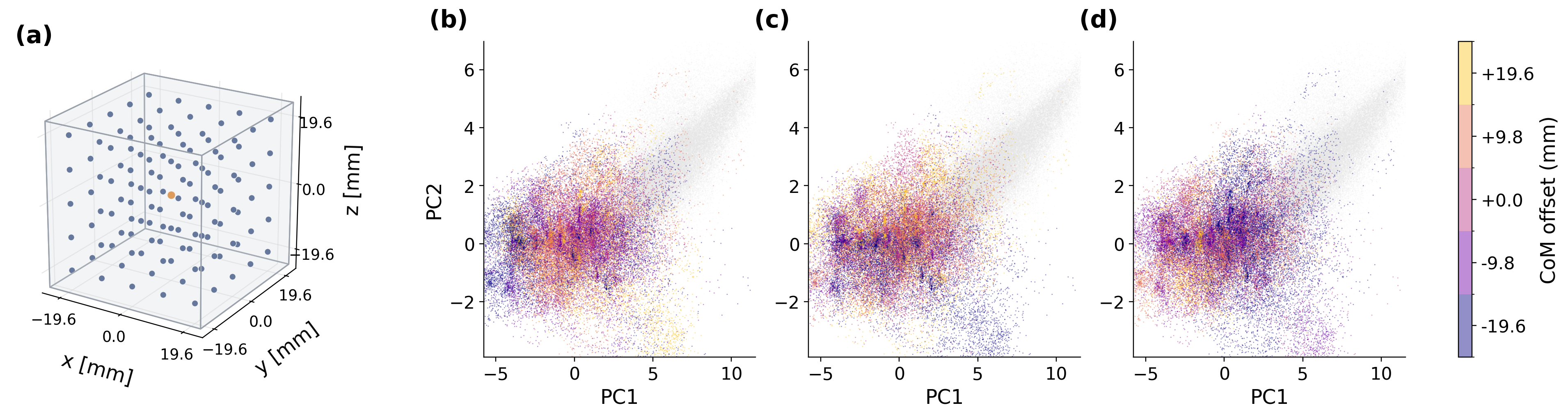}
  \caption{\textbf{(a)} Payload center-of-mass (CoM) variation in the training data.
    The CoM of the \SI{4.8}{\centi\metre} cube is offset independently along each
    axis to one of $\{0, \pm 9.8, \pm 19.6\}$\,\si{\milli\metre}, giving a
    $5^3 = 125$-point lattice (orange: cube center).
    \textbf{(b--d)} The latent $z$ of every training window, projected onto
    the same PC1--PC2 plane as Fig.~\ref{fig:qualitative_ic}(a) and coloured by the CoM
    offset along $\mathrm{x}$ (b), $\mathrm{y}$ (c) and $\mathrm{z}$ (d). Grey points
    are windows from the held-out \SIrange{2}{4}{\kilo\gram} payloads, whose CoM is
    fixed at the cube center.}
  \label{fig:com-latent}
\end{figure}

In the training set, the cube's CoM is drawn per episode from the lattice of
Fig.~\ref{fig:com-latent}(a), with uniform-cube inertia about the shifted
CoM. Since the cube is grasped at its center, this is also the CoM offset
from the gripper. It accounts for \(0.20\,\mathrm{N\,m}\) RMS of joint torque
on the validation set, about 5\% of the mass effect.

In the torque-intervention set, the CoM stays at the cube center and the
grasp point moves instead: the cube position given to the policy is shifted
along the hand \(x\)-axis (across the finger pads), with the target in-hand
offset drawn from \(\mathcal U(-1.6,1.6)\,\mathrm{cm}\) via a piecewise-linear
map calibrated in a pilot run. The realized offset (mean
\(-0.02\,\mathrm{cm}\), s.d.\ \(1.06\,\mathrm{cm}\)) is measured every step and
used only to convert predicted configurations to the cube pose. Other axes
cannot be varied: the gripper centers the cube along its closing axis, and
the table and fingers fix the grasp height.

\subsection{Payload Sweep under Figure-Eight Tracking}
\label{sec:app-payload-sweep}

The PPO policy lifts the cube to the center of a 3D figure-eight and tracks
\begin{equation}
    p^{\mathrm{goal}}(t) = p_c +
    \bigl(a_x\sin 2\theta,\;a_y\sin\theta,\;a_z\sin 2\theta\bigr),
    \qquad \theta=\frac{2\pi t}{T},
    \label{eq:app-figure-eight}
\end{equation}
with \(p_c=(0.53,0,0.37)\,\mathrm{m}\), \((a_x,a_y,a_z)=(7,15,12)\,\mathrm{cm}\),
\(T=3\,\mathrm{s}\), amplitude ramped in over 1\,s. Starting 4\,s in, the
mass rises linearly from \(0.1\) to \(4.0\,\mathrm{kg}\) over 20\,s, with
inertia updated every step and the CoM at the cube center. Of 192 episodes (32\,s each), 33 stay within
the slip limit; we evaluate all 32,967 windows during the ramp.
For Fig.~\ref{fig:qualitative_ic}(d), we additionally run the reverse schedule, in which the mass
decreases linearly from 4.0 to 0.1\,kg over the same interval.

\subsection{Torque Intervention}
\label{sec:app-torque-intervention}

Paired branches are recorded with the PPO pick-and-carry policy (seed 7000,
256 environments). Each environment draws an anchor step from control steps
52--72, during the moving carry. Branch~0 is the null run; every other branch
reruns the same seed, follows the null run up to the anchor and then replays
the null actions, so the policy cannot react. A \(\Delta\tau=0\) twin exactly
reproduces the null run. The intervention adds
\(\Delta\tau\in\{\pm10,\pm20,\pm30,\pm40\}\,\mathrm{N\,m}\) to joint 1, 2 or 3
for 15 steps; since the implicit PD drive partly cancels it, models receive
the measured \(\bar\tau\). Pairs are excluded if the null run hits a velocity
limit or the branch slips more than \(1\,\mathrm{cm}\), leaving 1,887 of
5,616 pairs (Section~\ref{sec:causal-dynamics-inference}).

\section{Simulator Dynamics and Evaluation References}
\label{sec:app-discrete-time-dynamics}


\paragraph{Mean acceleration and the velocity transition.} Each transition spans $\Delta t=20$\,ms and contains $S=40$ simulator
substeps. We use the interval-average acceleration
$\bar{\ddot q}_k=(\dot q_{k+1}-\dot q_k)/\Delta t
=S^{-1}\sum_{i=0}^{S-1}\ddot q_{k,i}$
as the dynamics target in Section~\ref{sec:contrastive}. Accordingly,
$\dot q_{k+1}=\dot q_k+\Delta t\,\bar{\ddot q}_k$.


\paragraph{Position integration and the remaining discretization discrepancy.} Models update position using the trapezoidal rule,
$q_{k+1}=q_k+\frac{\Delta t}{2}(\dot q_k+\dot q_{k+1})$,
whereas the simulator updates it at each substep,
$q_{k+1}^{\mathrm{sim}}=q_k+\frac{\Delta t}{S}\sum_{i=1}^{S}\dot q_{k,i}$.
Thus, the position updates may differ even with the true
interval-average acceleration. Cube poses are computed by forward
kinematics using the grasp transform measured at the last history frame.


\paragraph{Coupled torque balance for the arm and grasped object.} The measured arm torque includes the grasped object:
$\tau_{\mathrm{arm}}
=H_{\mathrm{arm}}\ddot q+c_{\mathrm{arm}}+g_{\mathrm{arm}}
+\tau_{\mathrm{obj}}$, where
$\tau_{\mathrm{obj}}=H_{\mathrm{obj}}\ddot q+c_{\mathrm{obj}}+g_{\mathrm{obj}}$.
The object terms are computed from its mass, center of mass, inertia,
and the arm Jacobians. We define
$H_{\mathrm{total}}=H_{\mathrm{arm}}+H_{\mathrm{obj}}$,
$c_{\mathrm{total}}=c_{\mathrm{arm}}+c_{\mathrm{obj}}$, and
$g_{\mathrm{total}}=g_{\mathrm{arm}}+g_{\mathrm{obj}}$.


\paragraph{Comparison of inertial, Coriolis and gravity torque contributions.} Simulator references are computed by averaging each substep contribution
over the interval:
$T_{H,k}^{\mathrm{sim}}=S^{-1}\sum_i H_{\mathrm{total},k,i}\ddot q_{k,i}$,
$T_{c,k}^{\mathrm{sim}}=S^{-1}\sum_i c_{\mathrm{total},k,i}$, and
$T_{g,k}^{\mathrm{sim}}=S^{-1}\sum_i g_{\mathrm{total},k,i}$.
We compare each predicted component with its reference and their sum
with the measured interval-average torque $\bar\tau_k$.

\end{document}